\documentclass[conference]{IEEEtran}

\usepackage{cite}
\usepackage{amsmath,amssymb,amsfonts}
\usepackage{algorithmic}
\usepackage{graphicx}
\usepackage{textcomp}
\usepackage{xcolor}
\def\BibTeX{{\rm B\kern-.05em{\sc i\kern-.025em b}\kern-.08em
    T\kern-.1667em\lower.7ex\hbox{E}\kern-.125emX}}

\usepackage{comment}
\usepackage[colorlinks=true, linkcolor=blue, citecolor=blue, urlcolor=black]{hyperref}
\usepackage{mathrsfs}
\usepackage{multirow}
\usepackage{makecell}
\usepackage{float}
\usepackage{booktabs}
\usepackage{color}

\begin{document}

\title{Learning from Limited and Incomplete Data:\\
A Multimodal Framework for Predicting Pathological Response in NSCLC}

\author{
\IEEEauthorblockN{
    Alice Natalina Caragliano\IEEEauthorrefmark{1}\IEEEauthorrefmark{5},
    Giulia Farina\IEEEauthorrefmark{1}\IEEEauthorrefmark{5},
    Fatih Aksu\IEEEauthorrefmark{1},
    Camillo Maria Caruso\IEEEauthorrefmark{1},
    Claudia Tacconi\IEEEauthorrefmark{2}, \\
    Carlo Greco\IEEEauthorrefmark{2}\IEEEauthorrefmark{3},
    Lorenzo Nibid\IEEEauthorrefmark{2}\IEEEauthorrefmark{3}, 
    Edy Ippolito\IEEEauthorrefmark{2}\IEEEauthorrefmark{3},
    Michele Fiore\IEEEauthorrefmark{2}\IEEEauthorrefmark{3},
    Giuseppe Perrone\IEEEauthorrefmark{2}\IEEEauthorrefmark{3}, \\
    Sara Ramella\IEEEauthorrefmark{2}\IEEEauthorrefmark{3},
    Paolo Soda\IEEEauthorrefmark{1}\IEEEauthorrefmark{4}, and
    Valerio Guarrasi\IEEEauthorrefmark{1}
}
\\[1.5ex]
\IEEEauthorblockA{\IEEEauthorrefmark{1}\textit{Unit of Artificial Intelligence and Computer Systems, Department of Engineering,}\\
\textit{Università Campus Bio-Medico di Roma, Rome, Italy}\\ 
Email: \{a.caragliano, valerio.guarrasi, p.soda\}@unicampus.it}
\IEEEauthorblockA{\IEEEauthorrefmark{2} \textit{Fondazione Policlinico Universitario Campus Bio-Medico, Rome, Italy}} 
\IEEEauthorblockA{\IEEEauthorrefmark{3}\textit{Department of Medicine and Surgery,}\\
\textit{Università Campus Bio-Medico di Roma, Rome, Italy}}
\IEEEauthorblockA{\IEEEauthorrefmark{4}\textit{Department of Diagnostics and Intervention, Radiation Physics, Biomedical Engineering,}\\ 
\textit{Umeå University, Umeå, Sweden}}
\IEEEauthorblockA{\IEEEauthorrefmark{5}\textit{These authors contributed equally to this work.}}
}

\maketitle

\begin{abstract}
Major pathological response (pR) following neoadjuvant therapy is a clinically meaningful endpoint in non-small cell lung cancer, strongly associated with improved survival. However, accurate preoperative prediction of pR remains challenging, particularly in real-world clinical settings characterized by limited data availability and incomplete clinical profiles. In this study, we propose a multimodal deep learning framework designed to address these constraints by integrating foundation model-based CT feature extraction with a missing-aware architecture for clinical variables. This approach enables robust learning from small cohorts while explicitly modeling missing clinical information, without relying on conventional imputation strategies. A weighted fusion mechanism is employed to leverage the complementary contributions of imaging and clinical modalities, yielding a multimodal model that consistently outperforms both unimodal imaging and clinical baselines. These findings underscore the added value of integrating heterogeneous data sources and highlight the potential of multimodal, missing-aware systems to support pR prediction under realistic clinical conditions. 

\end{abstract}


\section{Introduction}
Non-small cell lung cancer (NSCLC) represents the most common subtype of lung cancer, accounting for approximately 85\% of all diagnosed cases~\cite{cancernet2022}. Surgical resection remains the primary curative option for patients with early-stage and resectable locally advanced NSCLC. However, a substantial proportion of patients experience disease recurrence after surgery, underscoring the need for improved strategies to optimize treatment planning. In this context, neoadjuvant therapy (NAT) has gained increasing relevance as an effective approach to reduce tumor burden and eradicate micrometastatic disease prior to surgical resection, ultimately improving survival outcomes~\cite{betticher2006prognostic}. Major pathological response, defined as the presence of no more than 10\% viable tumor cells in resected specimens and hereafter referred to as pR, has emerged as a key indicator of tumor sensitivity to therapy, representing a clinically meaningful endpoint consistently associated with improved overall survival~\cite{provencio2020neoadjuvant}.  

Accurate preoperative prediction of pR would provide insights into NAT efficacy, supporting personalized treatment strategies and helping clinicians identify patients most likely to benefit from surgical intervention. In this context, computed tomography (CT) imaging plays a central role, as it is routinely acquired during clinical workflows and provides non-invasive information on tumor morphology and disease extent. State-of-the-art reports that radiomics features extracted from CT scans correlate with pR~\cite{liu2023development}; however, despite being non-invasive, radiomics relies on hand-crafted features that may not adequately capture the heterogeneity of tumor phenotypes. 

A powerful solution to address these issues is to leverage Deep Learning (DL), which has emerged as an effective tool for medical image analysis, enabling the automatic extraction of informative features directly from imaging data and demonstrating strong performance across a wide range of medical tasks~\cite{cellina2022artificial, caragliano2025exploring, mantegna2024benchmarking}. Recent studies have applied DL for predicting pR in NSCLC~\cite{qu2024non, caragliano2025doctor}, predominantly relying on unimodal imaging-based models. While promising, unimodal approaches may fail to fully capture the multifactorial nature of pR, as they do not exploit complementary information from different data sources.
In contrast, Multimodal DL (MDL) aims to overcome these limitations by integrating heterogeneous data modalities, enabling the construction of more comprehensive representations of tumor characteristics and patient status~\cite{guarrasi2025systematic}. MDL can thus be leveraged to reflect the complex determinants of treatment response by integrating imaging with clinical variables, such as tumor information, which provide complementary prognostic insights~\cite{caragliano2025multimodal}.  

However, the clinical translation of MDL approaches for pR prediction is hindered by two major challenges. First, medical datasets are often limited in size, increasing the risk of overfitting, particularly when training models from scratch on high-dimensional inputs. Second, real-world clinical data are inherently incomplete: missing variables are common due to heterogeneous acquisition protocols, retrospective data collection, human error during data entry, or data corruption. Conventional approaches typically address missing data through explicit imputation strategies, such as mean substitution or k-nearest neighbors. However, in sensitive domains such as oncology, imputing clinical variables may introduce biased or clinically implausible values, potentially distorting patient-specific information and affecting model reliability. 

Recent advances in foundation models (FMs) provide an effective solution to the challenge of limited data availability. Pretrained on large and diverse datasets, FMs are capable of extracting general, semantically rich representations that transfer effectively to downstream tasks~\cite{zhang2024challenges}. Leveraging FMs as imaging feature extractors enables robust representation learning from CT scans, even in small clinical cohorts~\cite{aksu2025nsclc}.

At the same time, robust integration of incomplete clinical data requires models explicitly designed to handle unavailable information. Missing-aware learning strategies, such as transformer-based architectures with adaptive masking mechanisms, allow models to operate directly on incomplete clinical profiles by selectively masking unavailable features, thereby avoiding the need for explicit data imputation~\cite{caruso2024not, maria}. 

To the best of our knowledge, no prior studies have proposed a multimodal method for pR prediction in NSCLC that jointly considers the challenges posed by limited data availability and incomplete clinical profiles. To address this gap, we propose a multimodal framework for pR prediction that combines FM-based feature extraction from CT scans with a missing-aware transformer architecture for clinical variables. CT and clinical data are then integrated, allowing each modality to contribute to the final prediction while preserving robustness to incomplete clinical data. Overall, our approach enables a non-invasive prediction of pR under realistic data constraints, supporting treatment planning in NSCLC.

The main contributions of this work are:
\begin{itemize}
    \item We propose a multimodal framework for pR prediction in NSCLC that integrates imaging and clinical modalities into a unified predictive model.
    \item We leverage a FM to extract robust and transferable imaging representations from CT scans, enabling effective learning in limited-size clinical cohorts.
    \item We adopt a missing-aware transformer-based architecture for modeling clinical data, allowing the proposed framework to operate directly on incomplete clinical profiles without explicit data imputation.  
\end{itemize}

\section{Methods}
Predicting pR following NAT in NSCLC requires integrating heterogeneous data sources while accounting for two key constraints of real-world clinical data: limited cohort size and incomplete clinical profiles. 
Training DL models from scratch on volumetric imaging data is particularly challenging in small clinical cohorts, as it increases the risk of overfitting. In addition, clinical data are frequently affected by missing values, which pose practical challenges for multimodal training. 

To address these limitations, we propose a multimodal framework that combines pretrained FM-based feature extraction from CT images with missing-aware modeling of clinical data, followed by a late fusion strategy. This modular design allows each modality to be processed using architectures tailored to its specific characteristics and data constraints.

\begin{figure*}[t] 
    \centering 
    \includegraphics[width=1\textwidth]{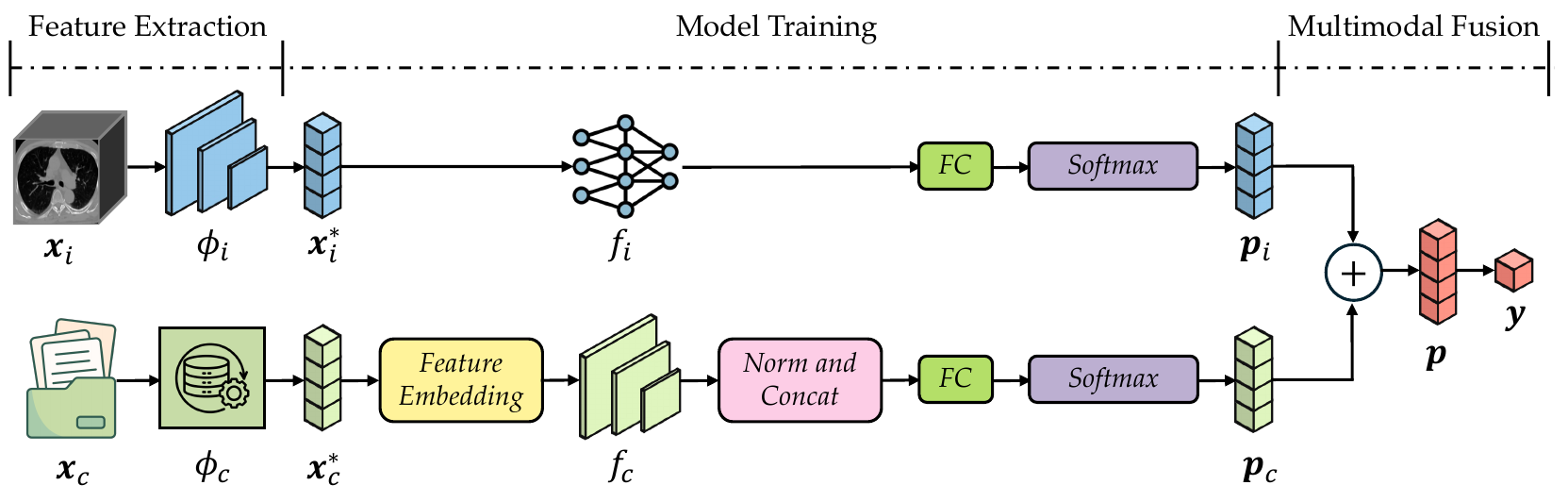} 
    \caption{{Overview of the proposed multimodal framework, organized into three main stages: feature extraction, model training, and multimodal fusion. Imaging and clinical modalities are processed through two parallel branches, and the resulting unimodal outputs are subsequently fused to predict pR.}}
    \label{fig:method} 
\end{figure*}

An overview of the proposed framework is illustrated in \hyperref[fig:method]{Fig.~\ref{fig:method}}. The architecture is organized into three main stages (\textit{feature extraction}, \textit{model training}, \textit{multimodal fusion}) and comprises two parallel branches corresponding to the \textit{imaging model} and the \textit{missing-aware clinical model}. Within these branches, the two modalities are handled separately during feature extraction and unimodal model training. Finally, their outputs are integrated within a multimodal fusion module to produce the final pR prediction.

\subsection{Feature extraction}
\paragraph{Imaging Model}
Let $\boldsymbol{X}_i \subseteq \mathbb{R}^{H \times W \times D}$ denote the space of volumetric CT images, where each scan $\boldsymbol{x}_i \in \boldsymbol{X}_i$ has spatial dimensions $H$ (height), $W$ (width), and $D$ (number of slices). To mitigate data scarcity, we decouple imaging representation learning from downstream classification by leveraging a pretrained FM for feature extraction. Specifically, we employ Merlin~\cite{blankemeier2024merlin}, a 3D Vision Transformer-based FM pretrained on large-scale multimodal datasets of paired CT volumes and radiology reports via a CLIP-style objective. This pretraining strategy enables Merlin to learn semantically rich and transferable anatomical representations that generalize across imaging tasks.
Volumetric CT features are extracted using the Merlin encoder, denoted as $\phi_i$, yielding the imaging representation:
\begin{equation}
\boldsymbol{x}_i^* = \phi_i(\boldsymbol{x}_i)
\end{equation}
where $\boldsymbol{x}_i^*$ represents the high-level feature vector associated with the input CT scan. 

\paragraph{Missing-Aware Clinical Model}
Let $\boldsymbol{X}_c \subseteq \mathbb{R}^{C}$ denote the  clinical input feature space, where each clinical feature vector $\boldsymbol{x}_c \in \boldsymbol{X}_c$ consists of $C$ variables. Prior to model training, clinical variables undergo a pre-processing pipeline to construct a standardized feature representation.
The resulting clinical feature vector is denoted as:
\begin{equation}
\boldsymbol{x}_c^{*} = \phi_c(\boldsymbol{x}_c)
\end{equation}
where $\phi_c$ represents the composite pre-processing function applied to the raw clinical data, as described in \hyperref[sec:dataset]{Section~\ref{sec:dataset}}.
Due to the nature of real-world clinical datasets, the pre-processed feature vector $\boldsymbol{x}_c^{*}$ may contain missing values, reflecting incomplete patient records. These missing values are explicitly accounted for in the subsequent modeling stage.

\subsection{Model training}

\paragraph{Imaging Model}
Following feature extraction, the imaging representation $\boldsymbol{x}_i^*$ is used as input to a unimodal classification model. Specifically, $\boldsymbol{x}_i^*$ is processed by a multilayer perceptron (MLP), denoted as $f_i$, followed by a fully connected (FC) layer and a softmax function to produce the predicted class probability vector $\boldsymbol{p}_i$. Finally, the imaging model is trained using the cross-entropy loss. 
By operating on FM-derived embeddings rather than raw voxel data, the unimodal imaging model remains lightweight and well suited for training under limited data availability.

\paragraph{Missing-Aware Clinical Model}

To robustly model structural missingness in clinical data, we employ the Not Another Imputation Method (NAIM)~\cite{caruso2024not}, a missing-aware transformer-based architecture explicitly designed to operate on incomplete feature vectors. Unlike standard DL models, which require complete inputs and hence rely on explicit imputation, NAIM introduces a paradigm shift by dynamically ignoring unavailable information rather than generating synthetic values.
To represent heterogeneous clinical variables, NAIM tokenizes each feature through a feature embedding module, which explicitly accounts for missing values.
Categorical variables are mapped using a learnable lookup table $\boldsymbol{E}^{\text{cat}}_j$: if a feature is observed, the corresponding trainable embedding vector is selected, while missing values are assigned a fixed padding embedding. The embedding of the $j$-th categorical feature is defined as:
\begin{equation}
\boldsymbol{e}^{\text{cat}}_j
= \boldsymbol{b}_j + \boldsymbol{E}^{\text{cat}}_j(x^{\text{cat}}_j)
\end{equation}
where $\boldsymbol{b}_j$ is a learnable bias term.
For numerical variables, NAIM preserves the magnitude of observed values by scaling the trainable embedding vector, selected from the lookup table $\boldsymbol{E}^{\text{num}}_j$, 
with the corresponding feature value. At the same time, the contribution of missing entries is suppressed by leveraging an indicator of feature availability $I_{\text{present}}$, which equals 1 when the feature is observed and 0 when it is missing. Accordingly, the embedding of the $j$-th numerical feature is defined as:
\begin{equation}
\boldsymbol{e}^{\text{num}}_j = \boldsymbol{b}_j + x^{\text{num}}_j \cdot \boldsymbol{E}^{\text{num}}_j(I_{\text{present}})
\end{equation}
The resulting feature embeddings form a sequence of tokens that is processed by the transformer encoder $f_{c}$ equipped with a missing-aware self-attention mechanism. To prevent missing features from influencing or being influenced by observed ones, NAIM employs an attention mask $\boldsymbol{M}$, defined such that $M_{ij} = -\infty$ when either feature $i$ or feature $j$ is missing, and $0$ otherwise. 
Let $\boldsymbol{Q}$, $\boldsymbol{K}$, and $\boldsymbol{V}$ denote the query, key, and value matrices derived from the embedded clinical tokens, the masked self-attention operation is defined as:
\begin{equation}
\scalebox{.98}{$
A(\boldsymbol{Q}, \boldsymbol{K}, \boldsymbol{V}) =
\text{ReLU}\!\left(
\text{Softmax}\!\left(\frac{\boldsymbol{Q}\boldsymbol{K}^{\text{T}}}{\sqrt{d_h}} + \boldsymbol{M}\right)
+ \boldsymbol{M}^{\text{T}}
\right)\boldsymbol{V}
$}
\end{equation}
where $d_h$ is the dimensionality of the attention head. 
By adding $-\infty$ to the attention scores prior to the softmax operation, the mask forces attention weights associated with missing features to exactly zero. The subsequent addition of the transposed mask $\boldsymbol{M}^{\text{T}}$ further ensures that any residual interactions involving missing features are explicitly suppressed. This double-sided masking guarantees that missing features neither contribute to nor receive information from the attention mechanism, ensuring that gradients propagate exclusively through observed data branches.
The output of the NAIM encoder is a contextualized clinical representation, which summarizes the available clinical information for a given patient. This representation is subsequently concatenated and passed to a FC layer followed by a softmax activation function, yielding the class probability vector $\boldsymbol{p}_c$. The clinical model is trained using the cross-entropy loss.

\subsection{Multimodal Fusion}

Since imaging and clinical modalities provide complementary information for pR prediction, the outputs of the unimodal imaging and clinical models are integrated through a late fusion strategy based on probability averaging. 

Let $\boldsymbol{p}_i$ denote the class probability vector predicted by the unimodal imaging model, and let $\boldsymbol{p}_c$ denote the class probability vector produced by the unimodal clinical model. The final multimodal prediction is obtained by computing the weighted element-wise average of the unimodal probability vectors, yielding the multimodal probability vector $\boldsymbol{p}$. Specifically, the following weighted averaging scheme is adopted:  

\begin{equation}
\boldsymbol{p} = \alpha \boldsymbol{p}_c + (1 - \alpha)\boldsymbol{p}_i, \label{late_fusion}
\end{equation}
where $\alpha \in [0,1]$ controls the relative contribution of the clinical and imaging modalities.
This late fusion strategy preserves the complementary contribution of the two modalities while maintaining robustness to missing clinical data. 


\section{Experimental setup}
To validate the proposed framework, we used an in-house dataset of NSCLC patients for the prediction of pR. 
\subsection{Dataset} 
The experiments were performed using a dataset of 100 NSCLC patients with TNM stage II-III~\cite{edition2017ajcc}, collected at \textit{Fondazione Policlinico Universitario Campus Bio-Medico of Rome}. All patients underwent neoadjuvant chemoradiation therapy followed by surgical resection. Among these patients, 36\% achieved a pR, defined as having no more than 10\% viable tumor cells in all specimens. 

This study was approved by two separate Ethical Committees. The retrospective phase was approved on October 30, 2012, and registered on ClinicalTrials.gov on July 12, 2018 (identifier: NCT03583723). The prospective phase was approved with the identifier 16/19 OSS. All patients provided written informed consent. 
Data used in this study are available from the corresponding author upon reasonable request.

For each patient, CT scans acquired within one month before the start of NAT were collected, along with a comprehensive set of clinical features capturing overall health status and treatment-related factors. These features encompass: (i) \textit{general patient characteristics}, including sex, age, weight, height, smoking status, number of cigarettes per day, family history of tumors, numeric pain rating scale, and comorbidities; (ii) \textit{tumor information}, including diagnosis, tumor stage, and TNM classification; (iii) \textit{biopsy details}, including diagnosis material, sampling site and technique; (iv) \textit{molecular biomarkers}, including EGFR, ALK, and PD-L1; (v) \textit{treatment details}, including induction chemotherapy scheme, number of induction chemotherapy cycles, chemotherapy scheme,  total and daily delivered radiation dose, number of radiotherapy sessions, radiotherapy technique and duration, days of treatment interruption, occurrence of permanent treatment interruption, and toxicities (esophageal, pulmonary, and hematologic).

Importantly, not all clinical variables were available for every patient, resulting in incomplete feature profiles. For some variables, such as family history of tumors and PD-L1 status, the proportion of missing values was particularly high (approximately 50\% of the cohort). Such levels of missingness make conventional imputation strategies potentially unreliable, as they may introduce biased estimates and distort the underlying data distribution. This scenario reflects realistic clinical practice and motivates the adoption of modeling strategies explicitly designed to account for missing information.

\subsection{Pre-processing} \label{sec:dataset} 

For the imaging modality, a standardized pre-processing procedure was applied prior to feature extraction to ensure compatibility with the Merlin encoding pipeline~\cite{blankemeier2024merlin}. First, all axial in-plane CT scans were resampled to a uniform resolution of $1.5 \times 1.5~\text{mm}^2$, while the slice thickness was resampled to $3.0~\text{mm}$ using bilinear interpolation, thereby correcting for inter-scan spacing variability. Second, voxel intensities were clipped using a standard lung window range (center = 0 HU, width = 2000 HU) and subsequently linearly normalized to the [0,1] range. Finally, the resampled volumes were cropped to a fixed spatial size of 224 $\times 224 \times 160$ voxels to ensure consistent input dimensions.

For the clinical modality, features were processed using a dedicated pipeline, in which categorical variables (e.g., comorbidities) were one-hot encoded to convert them into binary vectors, ordinal features (e.g., disease stage) were encoded based on their inherent ranking, and numerical features (e.g., patient age) were standardized using z-score normalization. 

\subsection{Experimental Configuration} 
To assess the effectiveness of the proposed framework, we conducted a series of experiments designed to analyze the individual and combined contributions of imaging and clinical data, as well as the impact of missing-aware clinical modeling. 

A detailed description of each experimental configuration is provided below.

\paragraph{Unimodal Imaging Model}
This experiment evaluates the unimodal imaging model trained on CT features extracted using the Merlin FM and subsequently processed by an MLP. This configuration assesses the predictive value of imaging information alone.

\paragraph{Unimodal Clinical Model}
This ablation study evaluates a unimodal clinical model based on a standard MLP, used in place of the NAIM architecture, to assess the impact of missing-aware clinical modeling. In this configuration, missing values are handled through conventional imputation strategies prior to model training. Specifically, numerical features are imputed using a k-nearest neighbors, while categorical variables are imputed using a most-frequent strategy. This pre-processing yields a fully complete clinical feature vector, which is then provided as input to the MLP. 

\paragraph{Unimodal Missing-Aware Clinical Model}
This experiment evaluates the unimodal clinical model trained on clinical variables using the NAIM architecture. By explicitly accounting for missing clinical features through a masked self-attention mechanism, this configuration provides insight into the predictive contribution of patient-specific clinical information under realistic conditions of incomplete data. 

\paragraph{Multimodal Model}
This experimental configuration evaluates the proposed multimodal framework based on late fusion of unimodal predictions. The final pR prediction is obtained by combining the class probability vectors produced by the unimodal imaging and clinical models through the weighted averaging scheme defined in \eqref{late_fusion}.
While $\alpha = 0$ corresponds to a purely imaging-based prediction and $\alpha = 1$ corresponds to a purely clinical-based prediction, intermediate values reflect different trade-offs between the two modalities. To systematically analyze the effect of modality balancing, $\alpha$ is varied from 0 to 1 with a step size of 0.1, enabling the evaluation of a continuum of fusion scenarios and the identification of configurations that optimally integrate imaging and clinical information.

\subsection{Training}
All experiments were conducted using consistent data splits to ensure a fair comparison across different configurations. The dataset was split into training (60\%), validation (20\%), and test (20\%) sets, via a 5-fold stratified cross-validation scheme. 

For the unimodal imaging model, the MLP classifier, consisting of two hidden layers, was optimized using Adam with an initial learning rate of $5 \times 10^{-4}$ and a weight decay of $1 \times 10^{-5}$. Training was performed for a maximum of 300 epochs, with early stopping applied based on the validation loss to prevent overfitting.

The unimodal clinical model based on the NAIM architecture was trained independently using the Adam optimizer. An initial learning rate of $1 \times 10^{-3}$ was employed, followed by an adaptive learning rate schedule that progressively reduced the learning rate by a factor of 10 whenever the validation loss did not improve for 25 consecutive epochs. The NAIM model was trained for a maximum of 1500 epochs, with early stopping based on the validation loss.


\begin{figure}[t] 
    \centering   
    \includegraphics[width=0.46\textwidth]{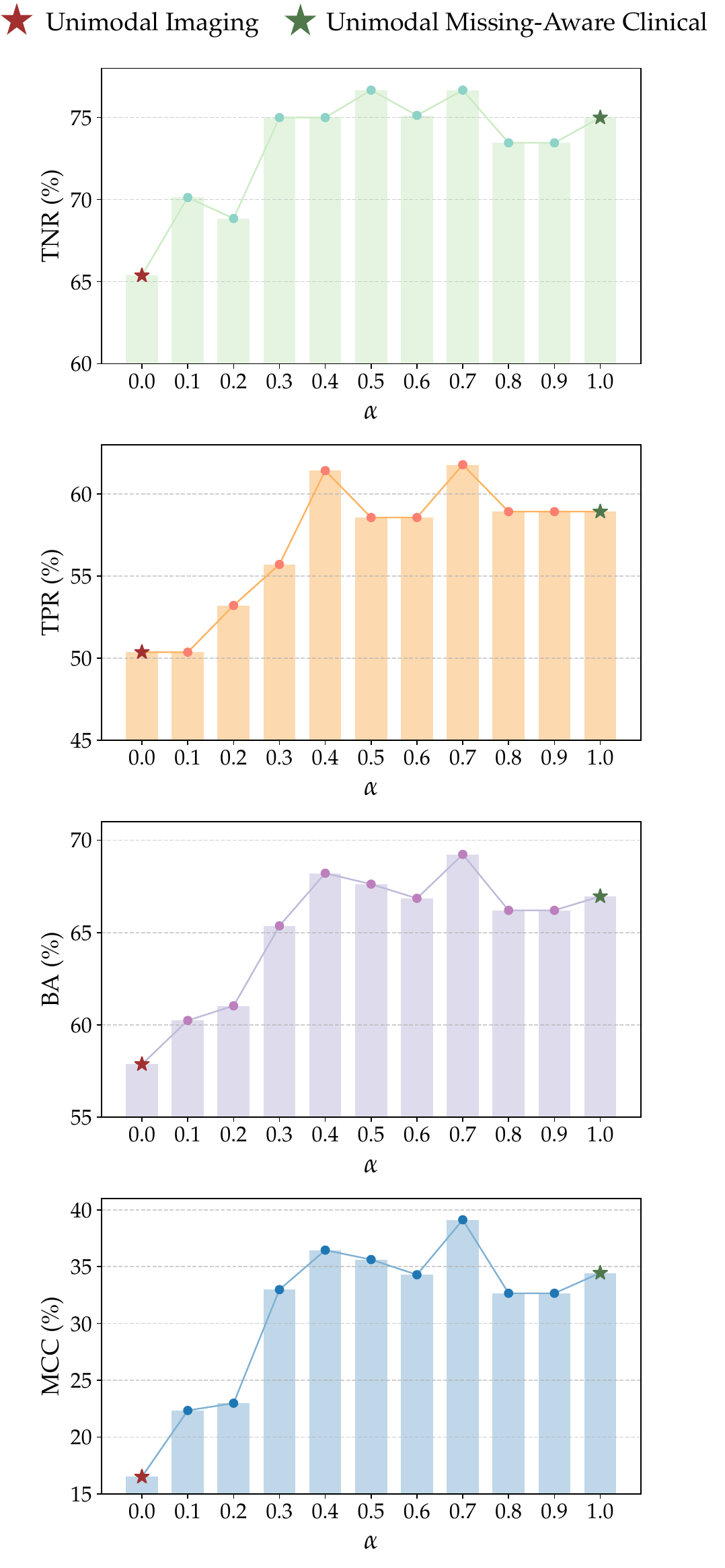} 
    \caption{{Effect of fusion weight $\alpha$ on model performance. Star markers indicate the unimodal imaging model ($\alpha=0$) and the unimodal missing-aware clinical model ($\alpha=1$). The highest performance is achieved at $\alpha=0.7$, underscoring the complementary contribution of clinical and imaging modalities.}} 
    \label{fig:plot} 
\end{figure}

\section{Results and discussion}

In this section, we present and discuss the results of the conducted experiments. Quantitative performance metrics are detailed in \hyperref[table:performance_results] {Table~\ref{table:performance_results}}, which reports True Positive Rate (TPR), True Negative Rate (TNR), Balanced Accuracy (BA), and Matthews Correlation Coefficient (MCC) across all experimental configurations. Results are reported as the mean values computed across the cross-validation folds, and for each metric the best-performing value is highlighted in bold.  Visual comparisons of performance metrics are provided in \hyperref[fig:plot]{Fig.~\ref{fig:plot}}.

\begin{table}[t]
\caption{Performance metrics under the different experimental configurations. For each metric, the best performing value is highlighted in bold.}
\label{table:performance_results}
\centering
\renewcommand{\arraystretch}{1.3}
\resizebox{\columnwidth}{!}{
\begin{tabular}{lccccc}
\toprule
\textbf{Experiment} & \textbf{TNR (\%)}  & \textbf{TPR (\%)} & \textbf{BA (\%)} & \textbf{MCC (\%)} \\ \midrule
\multirow{1}{*}{\textit{Unimodal Imaging}} & \text{65.38} & \text{50.36} & \text{57.87} & \text{16.50} \\ \midrule
\multirow{1}{*}{\textit{Unimodal Clinical}} & \text{74.87} & \text{52.50} &  \text{63.68} & \text{27.86} \\ \midrule                                   
\multirow{1}{*}{\textit{Unimodal Missing-Aware Clinical}} & \text{75.00} & \text{58.93} &  \text{66.96} & \text{34.45} \\ \midrule                                     
\multirow{1}{*}{\textit{Multimodal}}  & \textbf{76.67} & \textbf{61.79} &  \textbf{69.23} & \textbf{39.13} \\  \bottomrule

\end{tabular}
}
\end{table}

\subsection{Impact of Missing-Aware Modeling}
To assess the impact of explicitly handling missing clinical values, we compared NAIM against a clinical MLP baseline, where missing variables are handled using simple imputation techniques prior to model training. The NAIM-based clinical model outperformed the standard clinical model across all performance metrics (\hyperref[table:performance_results] {Table~\ref{table:performance_results}}), indicating the advantage of modeling missingness directly within the model architecture.

\subsection{Unimodal vs Multimodal}
As shown in \hyperref[table:performance_results] {Table~\ref{table:performance_results}}, the proposed multimodal framework based on weighted late fusion achieves the highest overall performance, outperforming both unimodal models across all evaluation metrics. The late fusion results reported in \hyperref[table:performance_results] {Table~\ref{table:performance_results}} correspond to the optimal fusion configuration with $\alpha = 0.7$, as confirmed by the fusion weight analysis in \hyperref[fig:plot]{Fig.~\ref{fig:plot}}.

The unimodal imaging model exhibits substantially lower performance, suggesting that imaging data alone has limited discriminative power for identifying patients achieving pR. In contrast, the missing-aware clinical model demonstrates stronger predictive capability, underscoring the relevance of clinical variables. The multimodal model achieves further performance gains over the clinical model alone, indicating that imaging features, although weaker in isolation, provide complementary information that enhances discrimination when appropriately integrated with clinical data.

\subsection{Impact of Fusion Weight}
\hyperref[fig:plot]{Fig.~\ref{fig:plot}} reports the performance trends obtained by varying the fusion weight $\alpha$ in \eqref{late_fusion},
thereby explicitly balancing the relative contribution of clinical and imaging modalities. $\alpha=0$ corresponds to the unimodal imaging model, which yields the lowest performance across metrics, whereas increasing $\alpha$ generally improves performance by progressively incorporating the stronger clinical signal.
The best overall trade-off is achieved at $\alpha=0.7$, where the model reaches its highest MCC and BA. This indicates that optimal performance is obtained when the clinical modality dominates the fusion, while the imaging modality still contributes meaningfully. Relying exclusively, or mostly exclusively, on clinical information ($\alpha>0.8$) does not maximize discrimination, highlighting the benefit of retaining a non-negligible imaging contribution. 
The same trend is observed for class-wise metrics: at $\alpha=0.7$, the model achieves the highest TPR values while maintaining a strong TNR, supporting a balanced performance between responders and non-responders. Overall, these results show that the proposed multimodal fusion provides an effective mechanism to combine modalities, with $\alpha=0.7$ representing the most effective balance in our experimental setting.

\section{Conclusion}
This study demonstrates the importance of combining imaging and clinical information for the prediction of pR in NSCLC patients undergoing NAT. By leveraging FM-based feature extraction from CT images and missing-aware modeling of clinical data, the proposed multimodal framework effectively addresses key challenges of limited data availability and incomplete clinical profiles. The results showed that late fusion of unimodal predictions provides an effective strategy for integrating complementary modalities, with optimal performance achieved when appropriately balancing clinical and imaging contributions. 
Future work could focus on validating the proposed framework on multicentric cohorts to further assess its robustness under heterogeneous clinical settings. Additionally, expanding our approach to other cancer subtypes could extend its broader potential in personalized oncology.

\section*{Acknowledgment}
Alice Natalina Caragliano is a Ph.D. student enrolled in the National Ph.D. in Artificial Intelligence, XXXIX cycle, course on Health and life sciences, organized by Università Campus Bio-Medico di Roma.
This work was partially funded by: 
i) PNRR 2022 MUR P2022P3CXJ-PICTURE (CUP C53D23009280001); 
ii) Cancerforskningsfonden Norrland project MP23-1122;
iii) Kempe Foundation project JCSMK24-0094.
Resources are provided by the \textit{National Academic Infrastructure for Supercomputing in Sweden} (NAISS) and the \textit{Swedish National Infrastructure for Computing} (SNIC) at Alvis @ C3SE. 

\bibliographystyle{IEEEtran}
\bibliography{bib_cbms}

@inproceedings{caragliano2025multimodal,
  title={{Multimodal Doctor-in-the-Loop: A Clinically-Guided Explainable Framework for Predicting Pathological Response in Non-Small Cell Lung Cancer}},
  author={Caragliano, Alice Natalina and others},
  booktitle={2025 International Joint Conference on Neural Networks (IJCNN)},
  pages={1--8},
  year={2025},
  organization={IEEE}
}

@article{caragliano2025doctor,
  title={{Doctor-in-the-Loop: An explainable, multi-view deep learning framework for predicting pathological response in non-small cell lung cancer}},
  author={Caragliano, Alice Natalina and others},
  journal={Image and Vision Computing},
  pages={},
  year={2025},
  publisher={Elsevier}
}

@article{guarrasi2025systematic,
  title={{A systematic review of intermediate fusion in multimodal deep learning for biomedical applications}},
  author={Guarrasi, Valerio and others},
  journal={Image and Vision Computing},
  pages={105509},
  year={2025},
  publisher={Elsevier}
}

@misc{cancernet2022,
    author       = "American Cancer Society",
    title        = "Key Statistics for Lung Cancer",
    year         = "",
    note         = "Available online at: \url{https://www.cancer.org/cancer/types/lung-cancer/about/key-statistics.html}"
}

@article{betticher2006prognostic,
  title={{Prognostic factors affecting long-term outcomes in patients with resected stage IIIA pN2 non-small-cell lung cancer: 5-year follow-up of a phase II study}},
  author={Betticher, DC and others},
  journal={British journal of cancer},
  volume={94},
  year={2006},
  publisher={Nature Publishing Group}
}

@article{provencio2020neoadjuvant,
  title={{Neoadjuvant chemotherapy and nivolumab in resectable non-small-cell lung cancer (NADIM): an open-label, multicentre, single-arm, phase 2 trial}},
  author={Provencio, Mariano and others},
  journal={The Lancet Oncology},
  volume={21},
  year={2020},
  publisher={Elsevier}
}

@article{cellina2022artificial,
  title={{Artificial intelligence in lung cancer imaging: unfolding the future}},
  author={Cellina, Michaela and others},
  journal={Diagnostics},
  volume={12},
  number={11},
  pages={2644},
  year={2022},
  publisher={MDPI}
}

@article{qu2024non,
  title={{Non-invasive prediction for pathologic complete response to neoadjuvant chemoimmunotherapy in lung cancer using CT-based deep learning: a multicenter study}},
  author={Qu, Wendong and others},
  journal={Frontiers in Immunology},
  volume={},
  pages={},
  year={2024},
  publisher={Frontiers Media SA}
}

@article{zhang2024challenges,
  title={{On the challenges and perspectives of foundation models for medical image analysis}},
  author={Zhang, Shaoting and Metaxas, Dimitris},
  journal={Medical image analysis},
  volume={91},
  pages={102996},
  year={2024},
  publisher={Elsevier}
}

@article{caruso2024not,
  title={{Not Another Imputation Method: A Transformer-based Model for Missing Values in Tabular Datasets}},
  author={Caruso, Camillo Maria and others},
  journal={arXiv preprint arXiv:2407.11540},
  year={2024}
}

@inproceedings{aksu2025nsclc,
  title={{NSCLC histological subtype classification from CT scans using generalist 3D medical foundation models}},
  author={Aksu, Fatih and others},
  booktitle={2025 IEEE 13th International Conference on Healthcare Informatics (ICHI)},
  pages={},
  year={2025},
  organization={}
}

@article{liu2023development,
  title={{Development and validation of a radiomics-based nomogram for predicting a major pathological response to neoadjuvant immunochemotherapy for patients with potentially resectable non-small cell lung cancer}},
  author={Liu, Chaoyuan and others},
  journal={Frontiers in Immunology},
  volume={14},
  pages={1115291},
  year={2023},
  publisher={Frontiers Media SA}
}

@article{blankemeier2024merlin,
  title={{Merlin: A vision language foundation model for 3d computed tomography}},
  author={Blankemeier, Louis and others},
  journal={Research Square},
  pages={rs--3},
  year={2024}
}

@article{edition2017ajcc,
  title={{}},
  author={Edition, S and others},
  journal={AJCC cancer staging manual},
  year={2017}
}

@article{maria,
    title={{MARIA: A multimodal transformer model for incomplete healthcare data}},
    author={Caruso, Camillo Maria and Guarrasi, Valerio and Soda, Paolo},
    journal={Computers in Biology and Medicine},
    volume={196},
    pages={110843},
    year={2025}
}

@article{caragliano2025exploring,
  title={{Exploring the Limitations of Virtual Contrast Prediction in Brain Tumor Imaging: A Study of Generalization Across Tumor Types and Patient Populations}},
  author={Caragliano, Alice Natalina and others},
  journal={NMR in Biomedicine},
  volume={},
  number={},
  pages={},
  year={2025},
  publisher={Wiley Online Library}
}

@inproceedings{mantegna2024benchmarking,
  title={{Benchmarking GAN-based vs classical data augmentation on biomedical images}},
  author={Mantegna, Massimiliano and others},
  booktitle={International Conference on Pattern Recognition},
  pages={92--104},
  year={2024},
  organization={Springer}
}
\end{document}